\documentclass{article}
\usepackage[utf8]{inputenc}
\pdfoutput=1
\usepackage[sc,osf]{mathpazo}
\usepackage[height=8.5in,width=6in,letterpaper]{geometry}
\usepackage[sort&compress,numbers]{natbib}
\usepackage[colorlinks=true,citecolor=blue,breaklinks]{hyperref}
\usepackage[hyphenbreaks]{breakurl} 
\usepackage[tracking]{microtype}

\makeatletter
\newlength\aftertitskip     \newlength\beforetitskip
\newlength\interauthorskip  \newlength\aftermaketitskip

\renewcommand\thefootnote{\textcolor{red}{\arabic{footnote}}}

\def\maketitle{\par
 \begingroup
   \def\thefootnote{\color{red}\fnsymbol{footnote}}
   \def\@makefnmark{\hbox to 4pt{$^{\@thefnmark}$\hss}}
   \@maketitle \@thanks
 \endgroup
\setcounter{footnote}{0}
 \let\maketitle\relax \let\@maketitle\relax
 \gdef\@thanks{}\gdef\@author{}\gdef\@title{}\let\thanks\relax}

\def\@startauthor{\noindent \normalsize\bf}
\def\@endauthor{}
\def\@starteditor{\noindent \small {\bf Editor:~}}
\def\@endeditor{\normalsize}
\def\@maketitle{\vbox{\hsize\textwidth
 \linewidth\hsize \vskip \beforetitskip
 {\begin{center} \LARGE\@title \par \end{center}} \vskip \aftertitskip
 {\def\and{\unskip\enspace{\rm and}\enspace}%
  \def\addr{\small\it}%
  \def\email{\hfill\small\tt}%
  \def\name{\normalsize\bf}%
  \def\AND{\@endauthor\rm\hss \vskip \interauthorskip \@startauthor}
  \@startauthor \@author \@endauthor}
}}

\makeatother

\pdfoutput=1                    

\usepackage{floatrow} 
\newfloatcommand{capbtabbox}{table}[][\FBwidth]

\usepackage{graphicx}
\usepackage{amsmath,amsthm,amssymb,bm} 
\usepackage{amsfonts}
\usepackage{mathrsfs}
\usepackage{caption}
\usepackage{subcaption}
\usepackage{multirow}
\usepackage{xspace}
\usepackage{array}
\usepackage{enumerate}
\usepackage{algorithm}
\usepackage{algorithmicx,algpseudocode}
\usepackage{hyperref}
\usepackage{stmaryrd}
\usepackage{appendix}
\usepackage{wrapfig}
\usepackage{hyperref}
\usepackage{sidecap}
\usepackage{booktabs}
\numberwithin{equation}{section}

\theoremstyle{plain}

\theoremstyle{definition}

\theoremstyle{remark}

\newcommand{\ours}{\text{DAN}\xspace}

\newcommand{\bp}{\mathbb{P}}

\newcommand{\single}{sample point\xspace}
\newcommand{\singles}{sample points\xspace}
\newcommand{\multi}{sample\xspace}
\newcommand{\multis}{samples\xspace}

\newcount\COMMENTs  
\usepackage{color}
\definecolor{darkgreen}{rgb}{0,0.5,0}
\definecolor{purple}{rgb}{1,0,1}
\newcommand{\comm}[2]{\ifnum\COMMENTs=1\textcolor{#1}{#2}\fi}

\title{Distributional Adversarial Networks}

\author{\name Chengtao Li \thanks{Authors contributed equally.} \email{ctli@mit.edu}\\ 
  \name David Alvarez-Melis \footnotemark[1]\email{dalvmel@mit.edu}\\
  \name Keyulu Xu \email{keyulu@mit.edu}\\
  \name Stefanie Jegelka \email{stefje@csail.mit.edu}\\
  \name Suvrit Sra \email{suvrit@mit.edu}\\
  \addr{Massachusetts Institute of Technology}
}

\begin{document}
\maketitle

\begin{abstract} 

	We propose a framework for adversarial training that relies on a \emph{\multi} rather than a single \emph{sample point} as the fundamental unit of discrimination. Inspired by discrepancy measures and two-sample tests between probability distributions, we propose two such distributional adversaries that operate and predict on samples, and show how they can be easily implemented on top of existing models. Various experimental results show that generators trained with our distributional adversaries are much more stable and are remarkably less prone to mode collapse than traditional models trained with pointwise prediction discriminators. The application of our framework to domain adaptation also results in considerable improvement over recent state-of-the-art.

\end{abstract} 

\section{Introduction}
\label{sec:intro}

Adversarial training of neural networks, especially Generative Adversarial Networks (GANs) \cite{goodfellow2014generative}, have proven to be a powerful tool for learning rich models, leading to outstanding results in various tasks such as realistic image generation, text to image synthesis, 3D object generation, and video prediction \cite{reed2016generative, 3dgan, vondrick2016generating}. Despite their success, GANs are known to be difficult to train. The generator and discriminator oscillate significantly from iteration to iteration, and slight imbalances in their capacity frequently causes the training to diverge. Another common problem suffered by GANs is \emph{mode collapse}, where the distribution learned by the generator concentrates on a few modes of the true data distribution, ignoring the rest of the space. In the case of images, this failure results in generated images that albeit realistic, lack diversity and reduce to a handful of prototypes.

There has been a flurry of research aimed at understanding and addressing the causes behind the instability and mode collapse of adversarially-trained models. The first insights came from \citet{goodfellow2014generative}, who noted that one of the leading causes of training instability was saturation of the discriminator. \citet{arjovsky2017towards} formalized this idea by showing that if the two distributions have supports that are disjoint or concentrated on low-dimensional manifolds that do not perfectly align, then there exists an optimal discriminator with perfect classification accuracy almost everywhere, and for which usual divergences (KL, Jensen-Shannon) will max-out. In follow-up work \cite{arjovsky2017wasserstein}, the authors propose an alternative training scheme based on an adversary that estimates the Wasserstein distance, instead of the Jensen-Shannon divergence, between real and generated distributions.

In this work, we take a novel approach at understanding the limitations of adversarial training, and propose a framework that brings the discriminator closer to a truly distributional adversary. Through the lens of comparisons of distributions, the objective of the original GAN can be understood as a Jensen-Shannon divergence test that uses a single \single\footnote{Throughout this paper we will use the term \emph{\multi} to refer to a set of instances generated from a distribution, \emph{\single} to refer to an element of that sample.} (disguised as batches, but operated on independently by the discriminator). We show that this approach is far from ideal from the statistical perspective and might partially explain the mode-collapsing behavior of GANs. Thus, we posit that it is desirable instead to consider discriminators that operate on \multis. There is a vast literature on two-sample tests and population comparisons from which one can draw inspiration. We propose methods based on two such approaches. Both of these methods operate on \multis and base their predictions upon their aggregated information. Furthermore, our approach can be understood as a variant of classical kernel two-sample tests, except that now the kernels are parametrized as deep neural networks and fully learned from data.

A very appealing characteristic of our framework is that it is surprisingly easy to impose on current methods. We show how off-the-shelf discriminator networks can be made \emph{distribution-aware} through simple modifications to their architecture, yielding an easy plug-and-play setup. We put our framework to test in various experimental settings, from benchmark image generation tasks to adversarial domain adaptation. Our experimental results show that this novel approach leads to more stable training, to remarkably higher performances, and most importantly, to significantly better mode coverage.\footnote{Code is available at \url{https://github.com/ChengtaoLi/dan}}

\paragraph{Contributions.} The main contributions of this work are as follows:\\[3pt]
$\bullet\;\;$ We introduce a new \emph{distributional} framework for adversarial training of neural networks, which operates on genuine \emph{\multis} rather than a \singles. \\
$\bullet\;\;$ We propose two types of adversarial networks based on this distributional approach, and show how existing models can seamlessly be adapted to fit within this framework. \\
$\bullet\;\;$ We empirically show that our distributional adversarial framework leads to more stable training and significantly better mode coverage compared to single-point-sample methods. The direct application of our framework to domain adaptation results in considerably higher performances over state-of-the-art.

\section{Distributional Approaches to Adversarial Training} 
\label{sec:distributional_approaches_to_adversarial_training}

\subsection{A case against single sample point discriminators}\label{sec:intuition} 
\label{sub:a_case_for_sample_based_discriminators}

To motivate our distributional approaches, we take the original GAN setting to illustrate intuitively how training with \single-based adversaries might lead to mode collapse in the generator. Recall the GAN objective function:
\begin{align}\label{eq:ganobj}
 \min_G \max_D \left\{\mathbb{E}_{x\sim \bp_x} [\log D(x)] + \mathbb{E}_{z\sim \bp_z } [\log (1- D(G(z)))] \right\}
\end{align}
where $D: \mathbb{R}^n \rightarrow [0, 1]$ maps a \single to the probability that it comes from data distribution $\bp_x$, and $G: \mathbb{R}^m \rightarrow \mathbb{R}^n$ maps a noise vector $z \in \mathbb{R}^m$, drawn from a simple distribution $\bp_z$, to the original data space. This in turn defines an implicit distribution $\bp_G$ for $G$'s generated outputs.

\begin{SCfigure}[0.6]
\centering
\includegraphics[width=0.53\textwidth]{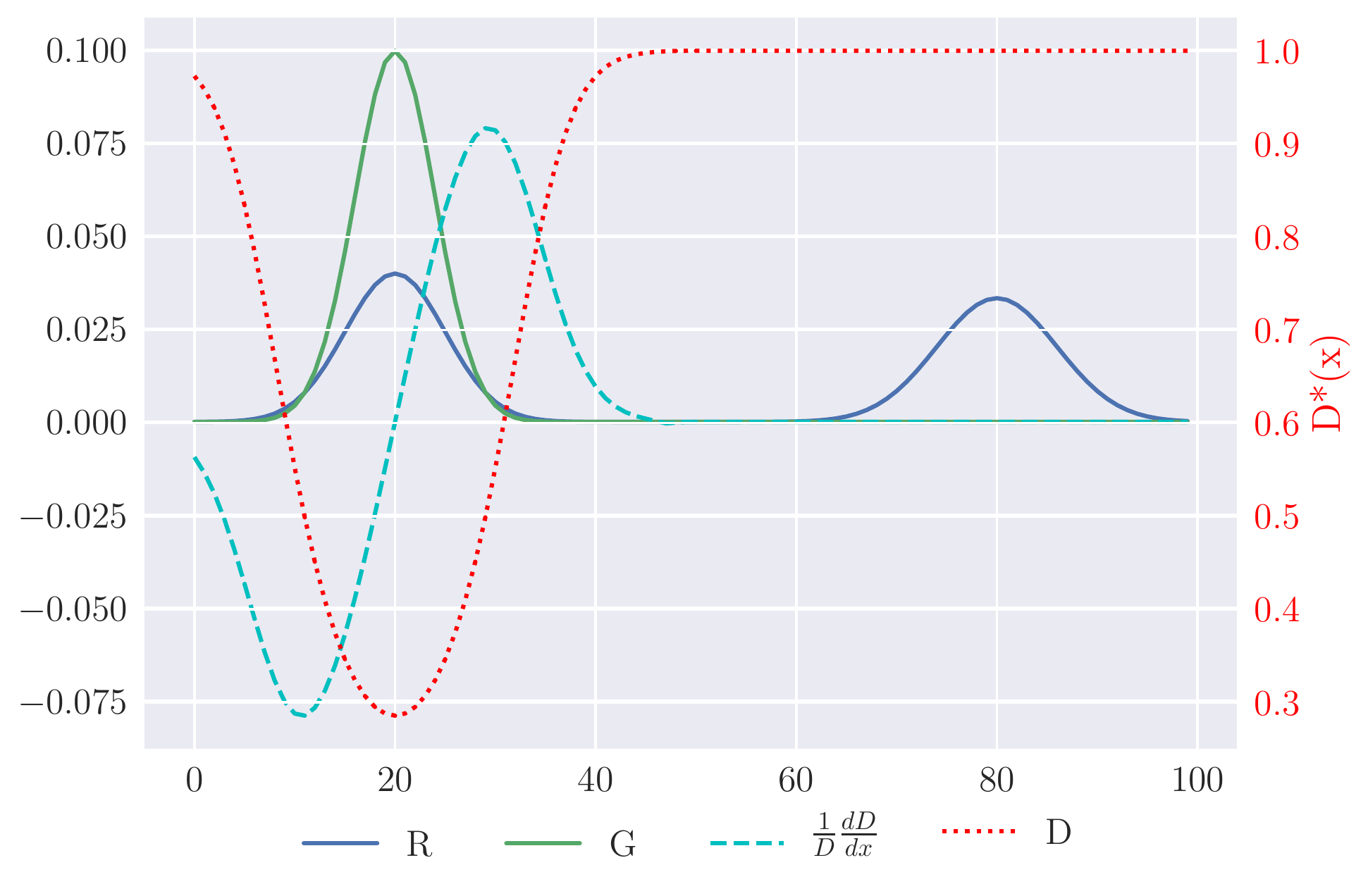}
\caption{Simple setting to explain the intuition behind mode-collapse behavior in \single discriminators with logistic loss. Gradients with respect to generated points $x$ are weighted by the term $-\frac{1}{D}\frac{dD}{dx}$ (cyan dashed line), so gradients corresponding to points close to the second mode will be dominated by those coming from the first mode.}\label{fig:intuition_importance}
\end{SCfigure}
To prevent loss saturation early in the training, G is trained to maximize $\log D(G(z))$ instead. As shown by ~\citet{goodfellow2014generative}, the discriminator converges in the limit to $D^*(x)= \bp_x(x)/(\bp_x(x)+\bp_G(x))$ for a fixed $G$. In this limiting scenario, it can be shown that given sample $Z = \{z^{(1)},\dots,z^{(B)}\}$ drawn from the noise distribution $\bp_z$, the gradient of G's loss with respect to its parameters $\theta_G$ is given by
\begin{equation}\label{eq:weights_gradients_gan}
 \nabla_{\theta_G}loss(Z) = {1\over B}\sum_{i=1}^B \frac{1}{D(G(z^{(i)}))} \nabla D(G(z^{(i)}))\left[\frac{d}{d\theta}G(z^{(i)})\right]
\end{equation}
where we slightly abuse the notation $\frac{d}{d\theta}G$ to denote $G's$ Jacobian matrix. Thus, the gradient with respect to each \single is weighted by terms of the form ${\nabla D(x_G^{(j)})\over D(x_G^{(j)})}$, where $x_G^{(j)}:=G(z^{(j)})$. These terms can be interpreted as relative slopes of the discriminator's confidence function. Their magnitude depends on the relation between $\frac{\partial D}{\partial x_G}$ (the slope of $D$ around $x_G$) and $D(x_G)$, the confidence on $x_G$ being drawn from $\bp_x$. Although the magnitude and sign of this ratio depend on the relation between these two opposing terms, there is one notable case in which it is unambiguously low: values of $x$ where the discriminator has high confidence of real samples \emph{and} flat slope. In other words, this weighting term will vanish in regions of the space where the generator's distribution has constant and low probability compared to the real distribution, such as neighborhoods of the support of $\bp_x$ where $\bp_G$ is missing a mode.  Figure~\ref{fig:intuition_importance} exemplifies this situation for a simple case in 1D where the real distribution is bimodal and the Generator's distribution is currently concentrated around one of the modes. The cyan dashed line corresponds to the weighting term $\nabla D(x_G)/D(x_G)$, confirming our analysis above that gradients for points around the second mode will vanish. 

The effect of this \emph{catastrophic averaging-out} of mode-seeking gradients during training is that it hampers G's ability to recover from mode collapse. Whenever $G$ does generate a point in a region where it misses a mode (which, by definition, already occurs with low probability), this example's gradient (which would update $G$'s parameters to move  mass to this region), will be heavily down-weighted and therefore dominated by high-weighted gradients of other examples in the batch, such as those in spiked high-density regions. At a high level, this phenomenon is a consequence of a myopic discriminator that bases its predictions on a single \single, leading to gradients that are not harmonized by global information. Discriminators that predict based on a whole \multi are a natural way to address this, as we will show in the next section.

\subsection{Distributional Adversaries}
\label{sec:da}

To mitigate the aforementioned failure case, we would like the discriminator to take a full \multi instead of a single \single into account for each prediction.  More concretely, we would like our discriminator to be a set function $M:2^{\mathbb{R}^d} \rightarrow \mathbb{R}$ that operates on a \multi $\{x^{(1)},\dots,x^{(n)}\}$ of potentially varying sizes. We construct the discriminator step-by-step as follows.

\paragraph{Deep Mean Encoder}\label{para:dme}

Despite its simple definition, the mean is a surprisingly useful statistic for discerning between distributions, and it is central to Maximum Mean Discrepancy (MMD)~\cite{gretton2005measuring,fukumizu2008kernel,smola2007hilbert}, a powerful discrepancy measure between distributions that enjoys strong theoretical guarantees.
Instead of designing a mapping explicitly, we propose to learn this function in a fully data-driven way, parametrizing $\phi$ as a neural network. Thus, a \emph{deep mean encoder} (DME) $\eta$ would have the form
\begin{align}
\eta(\bp) = \mathbb{E}_{x\sim\bp}[\phi(x)] 
\end{align}
In practice, $\eta$ only has access to $\bp$ through samples of finite size, thus effectively takes the form 
\begin{align}\label{eq:dme}
	\eta(\{x^{(1)},\dots,x^{(1)}\})  =  \frac{1}{n}\sum_{i=1}^n \phi(x^{(i)}) 
\end{align}
This distributional encoder forms the basis of our proposed adversarial learning framework. In what follows, we propose two alternative adversary models that build upon this encoder to discriminate between samples, and thus generate rich training signal for the generator.

\paragraph{Sample Classifier}\label{para:SC}

The most obvious way to use \eqref{eq:dme} as part of a discriminator in adversarial training is through a classifier. That is, given a vector encoding a \multi (drawn either from the data $\bp_x$ or generated distributions $\bp_G$), the classifier $\psi_S$ outputs $1$ to indicate the sample was drawn from $\bp_x$ and $0$ otherwise. The discrepancy between two distributions could then be quantified as the confidence of this classifier. Using a logistic loss, this yields the objective function
\begin{align}\label{eq:mb}
d_S(\bp_0, \bp_1) &= \log(\psi_S(\eta(\bp_1))) + \log(1 - \psi_S(\eta(\bp_0))),
\end{align}
This is analogous to the original GAN objective \eqref{eq:ganobj}, but differs from it in a crucial aspect: here the expectation is \emph{inside} of $\psi_S$. In other words, whereas in \eqref{eq:ganobj} the loss of a sample is defined as the expected loss of the sample points, here the sample loss is a single value, based on the deep mean embedding of its expected value. Therefore, \eqref{eq:mb} can be thought of an extension of \eqref{eq:ganobj} to non-trivial sample sizes. Henceforth, we use $M_{S}:=\psi_S \circ \eta$ to denote the full-model (mean encoder and classifier), and we refer to it as the \emph{sample classifier}.
 
\paragraph{Two-sample Discriminator}\label{para:2S}

The sample classifier proposed above has one potential drawback: the nonlinearities of the discriminator function are separable across $\bp_0$ and $\bp_1$, restricting the interactions between distributions to be additive. To address this potential issue, we propose to shift from a classification to a \emph{discrepancy} objective, that is, given two samples drawn independently, the two-sample discriminator must predict whether they were drawn from the same or different distributions. Concretely, given two encoded representations $\eta(\bp_0)$ and $\eta(\bp_1)$, the two-sample discriminator $\psi_{2S}$ uses the absolute difference of their deep mean encodings, and outputs $\psi_{2S}(|\eta(\bp_0) - \eta(\bp_1)|) \in [0,1]$, reflecting its confidence on whether the two samples were indeed drawn from the same distribution. Again with a binary classification criterion, the loss of this prediction would be given by 
\begin{align}\label{eq:mt}
d_{2S}(\bp_0, \bp_1) = &\llbracket \bp_0 = \bp_1\rrbracket\log(\psi_{2S}(|\eta(\bp_0) - \eta(\bp_1)|)) + \nonumber\\
&\llbracket \bp_0 \ne \bp_1\rrbracket(1 - \log(\psi_{2S}(|\eta(\bp_0) - \eta(\bp_1)|))
\end{align}
As before, we use $M_{2S}:=\psi_{2S} \circ | \eta(\cdot) - \eta(\cdot)|$ to denote the full two-sample discriminator model.

\section{Distributional Adversarial Network}\label{sec:train}

Now, we use the above distributional adversaries in a new training framework which we call \emph{Distributional Adversarial Network~(\ours)}. This framework can easily be combined with existing adversarial training algorithms by a simple modification in their adversaries. 
In this section, we examine in detail an example application of DAN to the generative adversarial setting. In the experiments section we provide an additional application to adversarial domain adaptation.

To integrate a GAN into the DAN framework, we simply add the distributional adversary: 
\begin{align}\label{eq:danobj}
\min_{G} \max_{D, M_\xi}& V(G,D,M_\xi) = \nonumber\\
&\lambda_1 \mathbb{E}_{x\sim\bp_x, z\sim\bp_z}[\log D(x) + \log(1 - D(G(z)))] + \lambda_2 d_\xi(\bp_x, \bp_G),
\end{align}
where $\xi\in\{S, 2S\}$ indicates whether we use the sample classifier or two-sample discriminator.

\paragraph{A Note on Gradient Weight Sharing} 
In the case of the sample classifier ($\xi = S$) and with only the distributional adversary ($\lambda_1= 0$ and $\lambda_2 > 0$ above), a derivation as in Section~\ref{sec:intuition} shows that
the gradient of a sample of $B$ points generated by G is 
\begin{equation}\label{eq:weights_gradients_set}
 \nabla_{\theta_G}loss(Z) = \frac{1}{\psi_S(\eta_B)}  \nabla \psi_S(\eta_B) \left (\frac{1}{B} \sum_{i=1}^B \nabla \psi_S(G(z^{(i)})) \left[\frac{d}{d\theta}G(z^{(i)})\right] \right)
\end{equation}
where we use $\eta_B := \eta(\{G(z^{(1)}),\dots,G(x^{(B)})\})$ for ease of notation. Note that, as opposed to \eqref{eq:weights_gradients_gan}, the gradient for each $z^{(i)}$ is weighted by the same left-most discriminator confidence term. This has the effect of sharing information across samples when updating gradients: whether a \multi (encoded as a vector $\eta_B$) can fool the discriminator or not will have an effect on every \single's gradient. In addition, this helps prevent the vanishing gradient for points generated in regions of mode collapse, since other sample points in the batch will come, with high probability, from regions of high density, and thus the weight vector that this \single gets, $\nabla \psi_S(\eta_B) $, will be less likely to be zero than in the single-sample-point discriminator case. Several interesting observations arise from this analysis. First, \eqref{eq:weights_gradients_set} suggests that the true power of this sample-based setting lies in choosing a discriminator $\psi_\xi$ that, through non-linearities, enforces interaction between the points in the sample. Second, the notion of sharing information across examples occurs also in batch normalization (BN) \citep{ioffe2015batch}, although the mechanism to achieve this and the underlying motivation for doing it are different. This connection, which might provide additional theoretical explanation for BN's empirical success, is supported by our experimental results, in which we show that the improvements brought by our framework are much more pronounced when BN is not used. While the analysis here is certainly not a rigorous one, the intuitive justification for sample-based aggregation is clear, and is confirmed by our experimental results.

\begin{figure}[h!]
\centering
\includegraphics[width=\textwidth]{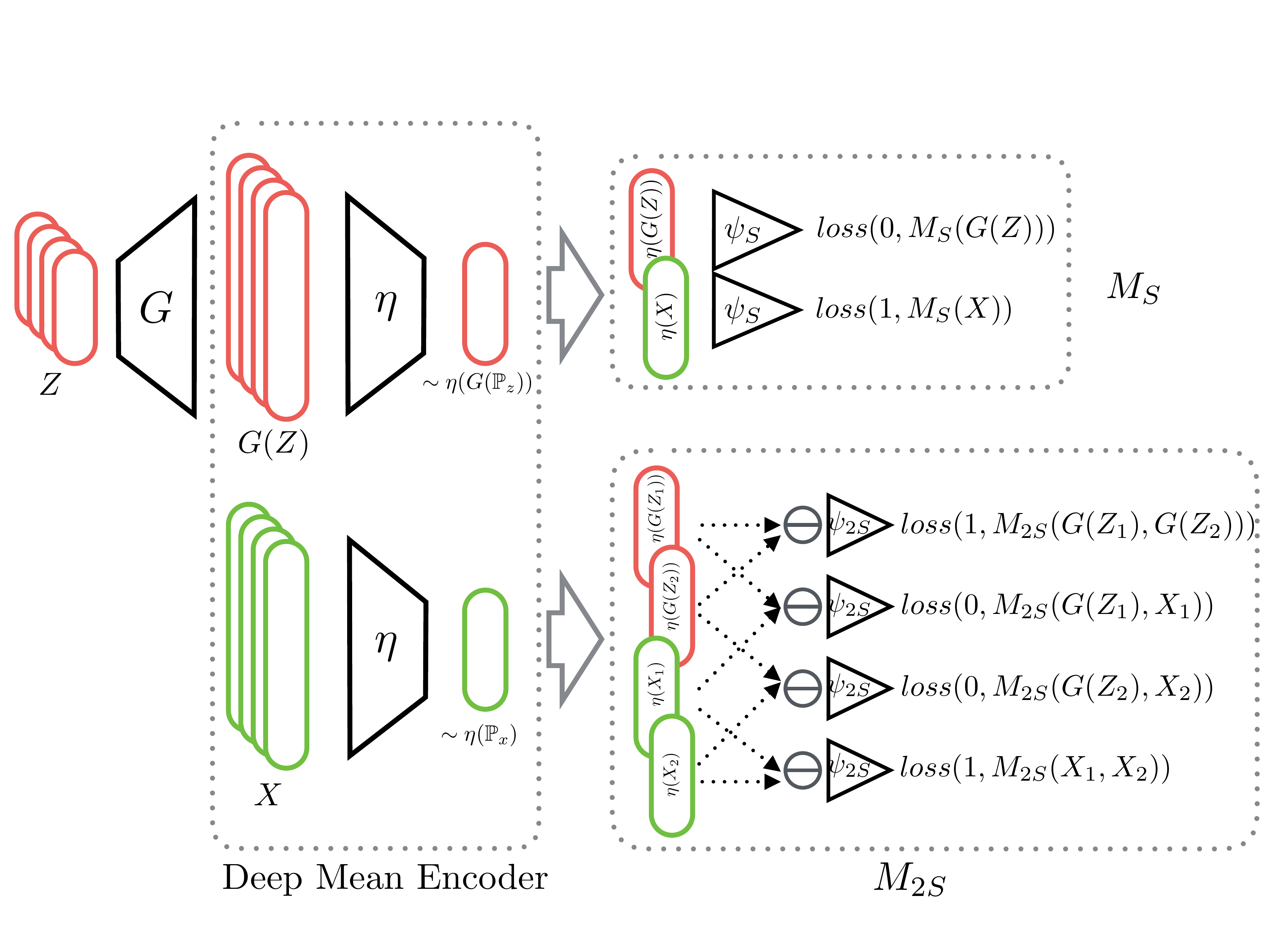}
\caption{\ours-S and \ours-2S models and corresponding losses, where $X=\{x^{(i)}\}_{i=1}^B\sim\bp_x$, $Z=\{z^{(i)}\}_{i=1}^B\sim\bp_z$, $X_1 :=\{x^{(i)}\}_{i=1}^{B\over 2}$, $X_2=\{x^{(i)}\}_{i={B\over 2}+1}^B$, $Z_1 :=\{z^{(i)}\}_{i=1}^{B\over 2}$ and $Z_2=\{z^{(i)}\}_{i={B\over 2}+1}^B$}
\label{fig:train}
\end{figure}

\paragraph{Training}
During training, all expectations of over data/noise distributions will be approximated via finite-sample averages. In each training iteration, we use minibatches (i.e.~\multis) from the data/noise distributions. We train by alternating between updates of the generator and adversaries.

While for \ours-S the training procedure is similar to that of GAN, \ours-2S requires a modified training scheme. Due to the form of the two-sample discriminator, we want a balanced exposure to pairs of samples drawn from the same and different distributions. Thus, every time we update $M_{2S}$, we draw minibatches $X = \{x^{(1)},\dots,x^{(B)}\} \sim \bp_x$ and $Z = \{z^{(1)},\dots,z^{(B)}\} \sim \bp_z$ from data and noise distributions. We then split each sample into two parts, $X_1 :=\{x^{(i)}\}_{i=1}^{B\over 2}, X_2=\{x^{(i)}\}_{i={B\over 2}+1}^B$, $Z_1 :=\{z^{(i)}\}_{i=1}^{B\over 2}, Z_2=\{z^{(i)}\}_{i={B\over 2}+1}^B$ and use the discriminator $M_{2S}$ to predict on each pair of $(X_1,G(Z_2))$, $(G(Z_1),X_2)$, $(X_1,X_2)$ and $(G(Z_1),G(Z_2))$ with target outputs $0$, $0$, $1$ and $1$, respectively. 
A visualization of this procedure is shown in Figure~\ref{fig:train} and the detailed training procedure is described in Appendix~\ref{app:sec:train}.

\subsection{Related Work} \label{sec:relation}

\paragraph{Discrepancy Measures}

The distributional adversary bears close resemblance to two-sample tests~\cite{lehmann2006testing}, where the model takes two samples drawn from potentially distinct distributions as input and produces a discrepancy value quantifying how different two distributions are.
A popular kernel-based variant is the \textit{Maximum Mean Discrepancy} (MMD)~\cite{gretton2005measuring,fukumizu2008kernel,smola2007hilbert}: 
\begin{small}
\begin{align*}
\text{MMD}^2(U,V) &= \|\tfrac1n \sum\nolimits_{i=1}^n \phi(u_i) - \tfrac1m\sum\nolimits_{j=1}^m \phi(v_j)\|_2^2
\end{align*}
\end{small}
where $\phi(\cdot)$ is some feature mapping, and $k(u,v) = \phi(u)^\top \phi(v)$ is the corresponding kernel function. An identity function for $\phi$ corresponds to computing the distance of the sample means.  More complex kernels result in distances of higher-order statistics between two samples. 

In adversarial models, MMD and its variants~(e.g., central moment discrepancy~(CMD)) have been used either as a direct objective for the target model or as a form of adversary~\cite{li2015generative, dziugaite2015training, sutherland2016generative, zellinger2017central,zellinger2017central}. However, such discrepancy measure requires hand-picking the kernel. 
An adaptive feature function, in contrast, may be able to adapt to the given distributions and thereby
 discriminate better.

Our distributional adversary tackles the above drawbacks and, at the same time, generalizes many discrepancy measures given that a neural network is a universal approximator. Since it is trainable, it can evolve as training proceeds. Specifically, it can be of simpler form when target model is weak at the beginning of training, and becomes more complex as target model becomes stronger.

\paragraph{Minibatch Discrimination}

Another relevant line of work involves training generative models by looking at statistics at minibatch level, known as \emph{minibatch discrimination}, initially developed in \cite{salimans2016improved} to stabilize GAN training. 
Batch normalization~\cite{ioffe2015batch} can be seen as performing some form of minibatch discrimination and it has been shown helpful for GAN training~\cite{radford2015unsupervised}. ~\citet{zhao2016energy} proposed a repelling regularizer that operates on a minibatch and orthogonalizes the pairwise sample representation, keeping the model from concentrating on only a few modes. 

Our distributional adversaries can be seen as learning a minibatch discriminator that adapts as the target model gets stronger. It not only enjoys the benefits of minibatch discrimination, leading to better mode coverage and stabler training (as shall be seem in Section~\ref{sec:exp}), but also eliminates the need of hand-crafting various forms of minibatch discrimination objectives.

\paragraph{Permutation Invariant Networks}

The deep mean encoder takes an (unordered) \multi to produce a mean encoding. A network whose output is invariant to the order of inputs is called \emph{permutation invariant network}. Formally, let $f$ be the network mapping, we have
$f(x_1,x_2,\ldots, x_n) = f(x_{\sigma(1)}, x_{\sigma(2)},\ldots, x_{\sigma(n)})$, 
where $n$ could be of varying sizes and $\sigma(\cdot)$ is any permutation over $[n]$. 

The problem of dealing with set inputs is being studied very recently.~\citet{vinyals2015order} consider dealing with unordered variable-length inputs with content attention mechanism. In later work of~\cite{ravanbakhsh2016deep, zaheer2016deep}, the authors consider the way of first embedding all samples into a fixed-dimensional latent space, and then adding them up to form a vector of the same dimension, which is further fed into another neural network. In~\cite{liu2017quality}, the authors proposed a similar network for embedding a set of images into a latent space. They use a weighted summation over latent vectors, where the weights are learned through another network.
The structure of our deep mean encoder resembles these networks in that it is permutation-invariant, but differs in its motivation---mean discrepancy measures---as well as its usage within discriminators in adversarial training settings.

\paragraph{Other Related Work}

Extensive work has been devoted to resolving the instability and mode collapse problems in GANs. One common approach is to train with more complex network architecture 
or better-behaving objectives~\cite{radford2015unsupervised,huang2017stacked,zhang2016stackgan,che2017mode,zhao2016energy,arjovsky2017wasserstein}. Another common approach is to add more discriminators or generators~\cite{durugkar2017generative,tolstikhin2017adagan} in the hope that training signals from multiple sources could lead to more stable training and better coverage of modes.


\section{Empirical Results}
\label{sec:exp}

We demonstrate the effectiveness of \ours training by applying it to generative models and domain adaptation.\footnote{Please refer to the appendix for all details on dataset, network architecture and training procedures.} In generative models, we observe remarkably better mode recovery than non-distributional models on synthetic and real datasets, through both qualitative and quantitative evaluation of generated sample diversity. In domain adaptation we leverage distributional adversaries to align latent spaces of source and target domains, yielding considerable improvements over state-of-the-art. 

\subsection{Synthetic Data: Mode Recovery in Multimodal Distributions}

\begin{figure}
\centering
\includegraphics[width=0.8\textwidth]{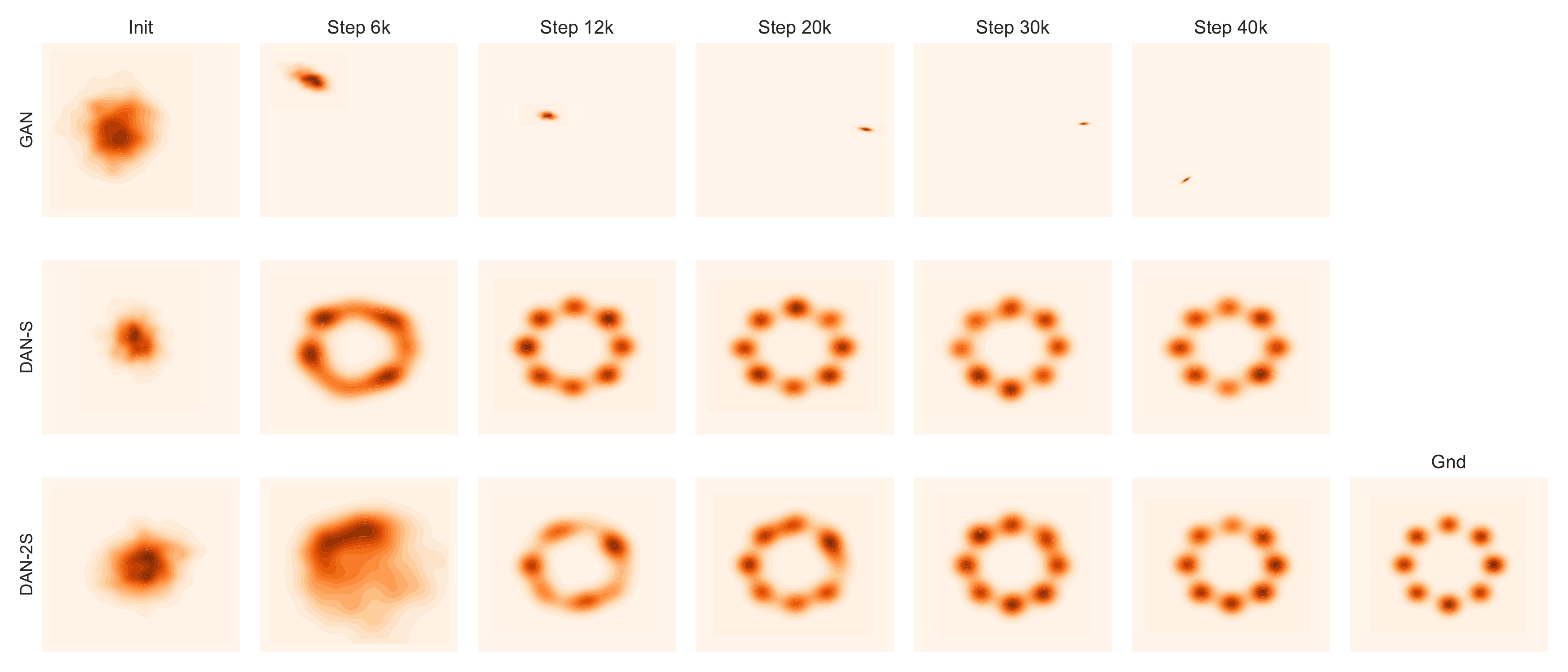}
\caption{Results for mode recovery on data generated from 8 Gaussian mixtures. The rightmost distribution is the true data distribution. While with GAN training the generator is only able to capture 1 of 8 modes, in both \ours-S and \ours-2S training we are able to recover all 8 modes. }
\label{fig:toy}
\end{figure}

We first apply DAN to generative models, where the data is a mixture of $8$ two-dimensional Gaussian distributions, with means aligned on a circle (bottom right of Figure~\ref{fig:toy}). The goal is for the generator to recover 8 modes on the circle. A similar task has been considered as a proof of concept for mode recovery in~\cite{metz2017unrolled,che2017mode}. 
All architectures are simple feed-forward networks with \texttt{ReLU} activations. To test the power of mode-recovery for DAN, we set $\lambda_1=0$ and $\lambda_2=1$, so only the distributional adversary is included in training. The results are shown in Figure~\ref{fig:toy}.

While the GAN generator produces samples oscillating between different modes, our distributional methods consistently and stably recover all eight modes of the true data distribution. Among these two, \ours-2S takes slightly longer to converge. We observe this multi-modal seeking behavior from early in the process, where the generator first distributes mass broadly covering all modes, and then subsequently sharpens the generated distribution in later iterations.

Note that here we set the architecture of \ours to be the same as GAN, with the only difference that the DAN adversary takes one extra step of averaging latent representations in the middle, which does not increase the number of parameters. Thus, DAN achieves a significantly better mode recovery but with the same number of network parameters.

\subsection{MNIST: Recovering Mode Frequencies}

Mode recovery entails not only \emph{capturing} a mode, i.e., generating samples that lie in the corresponding mode, but also \emph{recovering} the true probability mass of the mode. Next, we evaluate our model on this criterion. 
To do so, we train \ours on MNIST, which has a 10-class balanced distribution over digits, and compare the frequencies of generated classes against this target uniform distribution. Since the generated samples are unlabeled, we train an external classifier on MNIST to label the generated data.

Besides the original GAN, we also compared to two recently proposed generative models: RegGAN~\cite{che2017mode} and EBGAN~\cite{zhao2016energy}. To keep the approaches comparable, we use a similar neural network architecture in all cases, which we did not tailor to any particular model. We trained models without Batch Normalization (BN), except for RegGAN and EBGAN, which we observed to consistently benefit from it. We found that in tasks beyond simple data generation in low-dimensional spaces, it is beneficial to set $\lambda_1, \lambda_2 > 0$ for DAN. In this case, the generator will receive training signals from generated samples \emph{both locally and globally}, since both local~(single-sample) and distributional adversaries are used. 
Throughout the experiment we set $\lambda_1 = 1$ and $\lambda_2 = 0.2$. 

\begin{figure}
\centering
	\includegraphics[width=\textwidth]{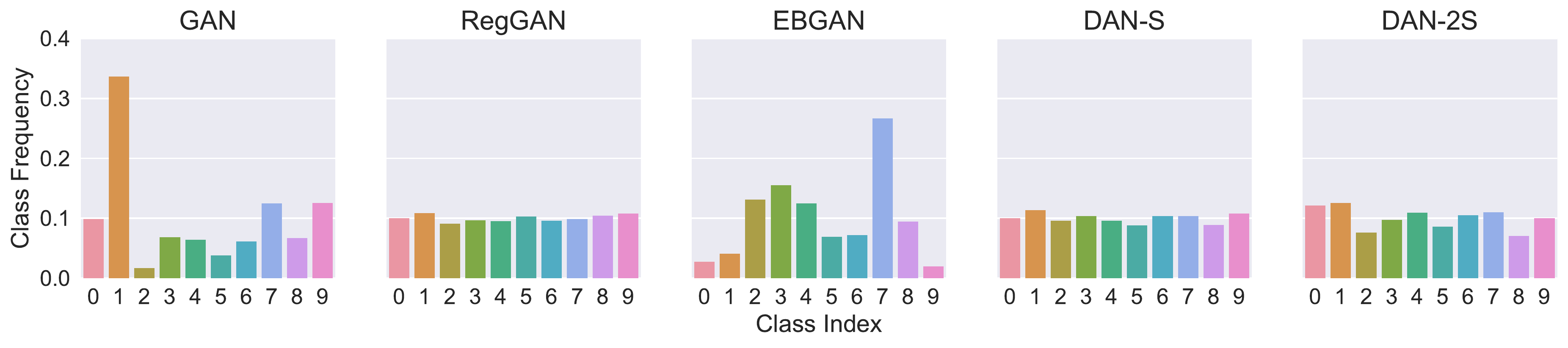}
	\caption{Class distribution of samples generated by various models on MNIST. Note that we showcase the \emph{best} one out of 10 random runs for GAN, RegGAN and EBGAN in terms of distribution entropy to give them an unfair advantage. For \ours we simply showcase a random run (the variance of the performance in FIgure~\ref{fig:mnist_stat} indicates that \ours-$\xi$ is stable across all runs). The best runs for RegGAN and EBGN also recover mode frequencies to some extent, but the performance varies a lot across different runs as in Figure~\ref{fig:mnist_stat}.}
\label{fig:mnist}
\end{figure}

\begin{SCfigure}[0.9]
\centering
	\includegraphics[width=0.46\textwidth]{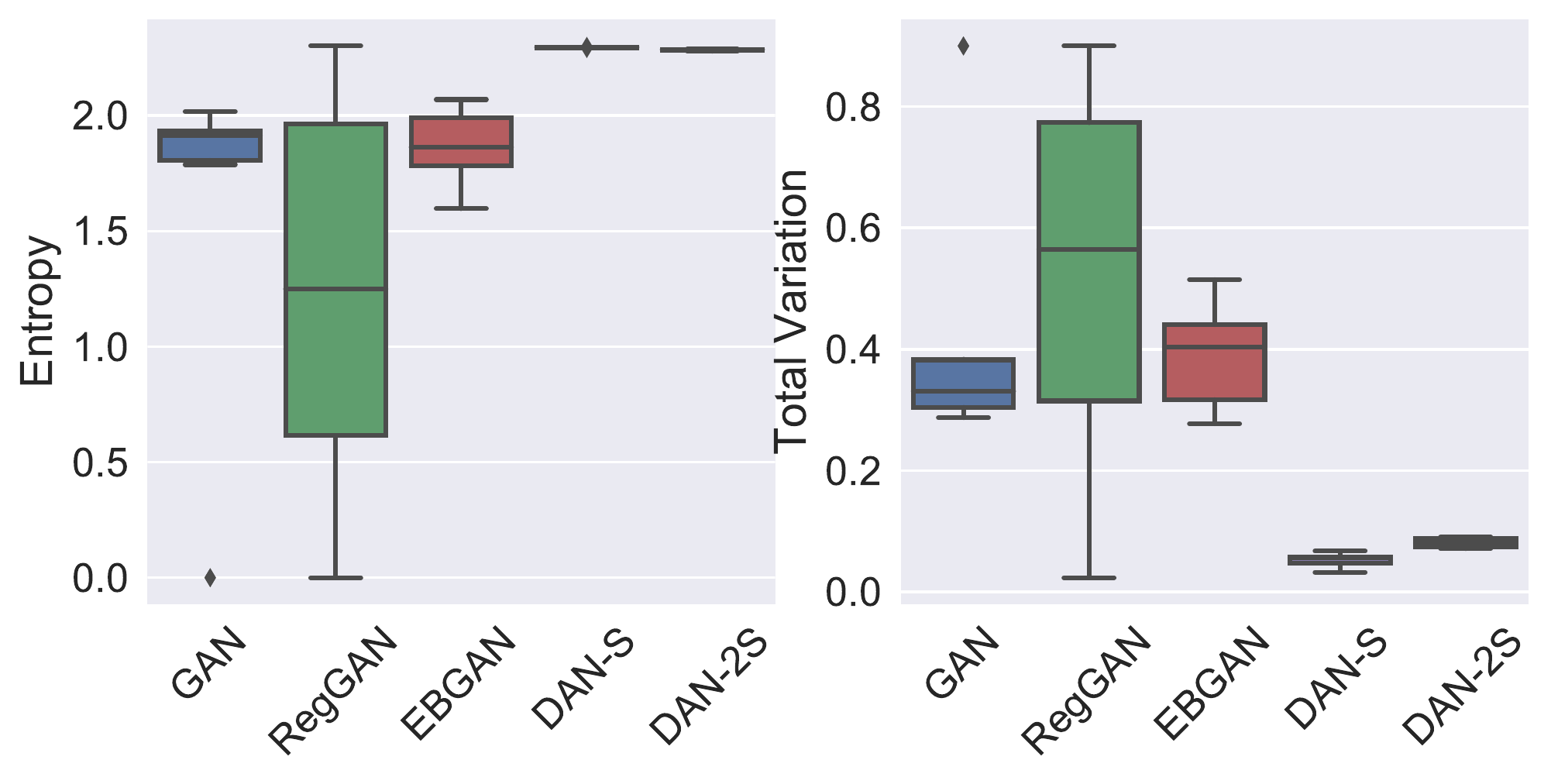}
	\caption{Performances of mode frequency recovery under 2 different measures: entropy of generated mode distribution and total variation distances between generated mode distribution and uniform one. \ours achieves the best and most stable mode frequency recovery.}
	\label{fig:mnist_stat}
\end{SCfigure}

 We show the results in Figure~\ref{fig:mnist} and~\ref{fig:mnist_stat}. Training with the original GAN leads to generators that place too much mass on some modes and ignore some others. While RegGAN and EBGAN sometimes generate more uniform distributions (such as the best-of-ten shown in Figure~\ref{fig:mnist}), this varies significantly across repetitions (c.f. Figure~\ref{fig:mnist_stat}), an indication of instability. \ours, on the other hand, \emph{consistently} recovers the true frequencies across repetitions.

\subsection{Image Generation: Sample Diversity as Evidence of Mode Coverage} 
\label{sub:image_generation}

\begin{figure}[h!]
\centering

\begin{subfigure}{0.23\textwidth}
	\includegraphics[width=\textwidth]{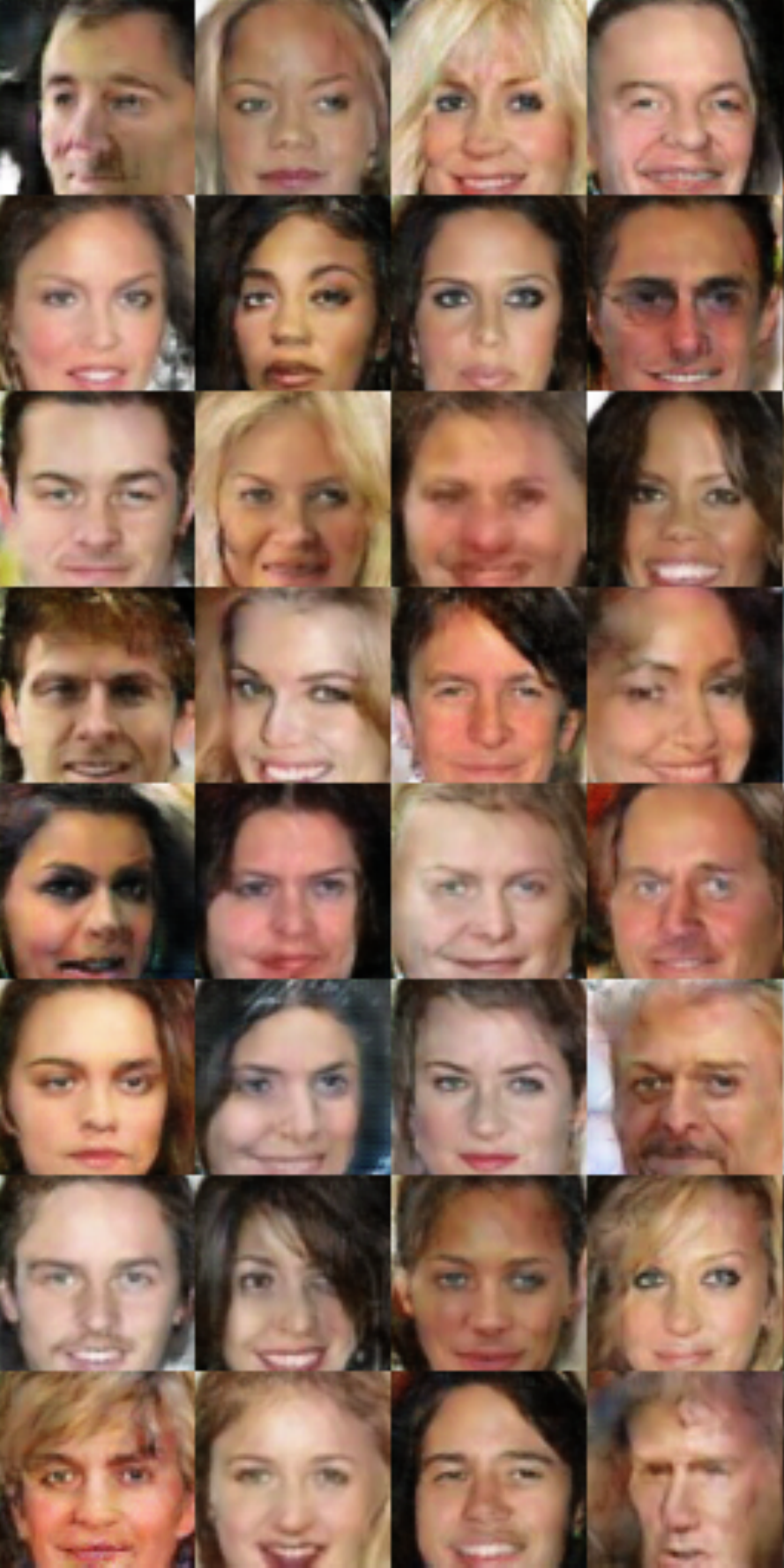}
	\caption{DCGAN, without BN}
\end{subfigure}	~
\begin{subfigure}{0.23\textwidth}
	\includegraphics[width=\textwidth]{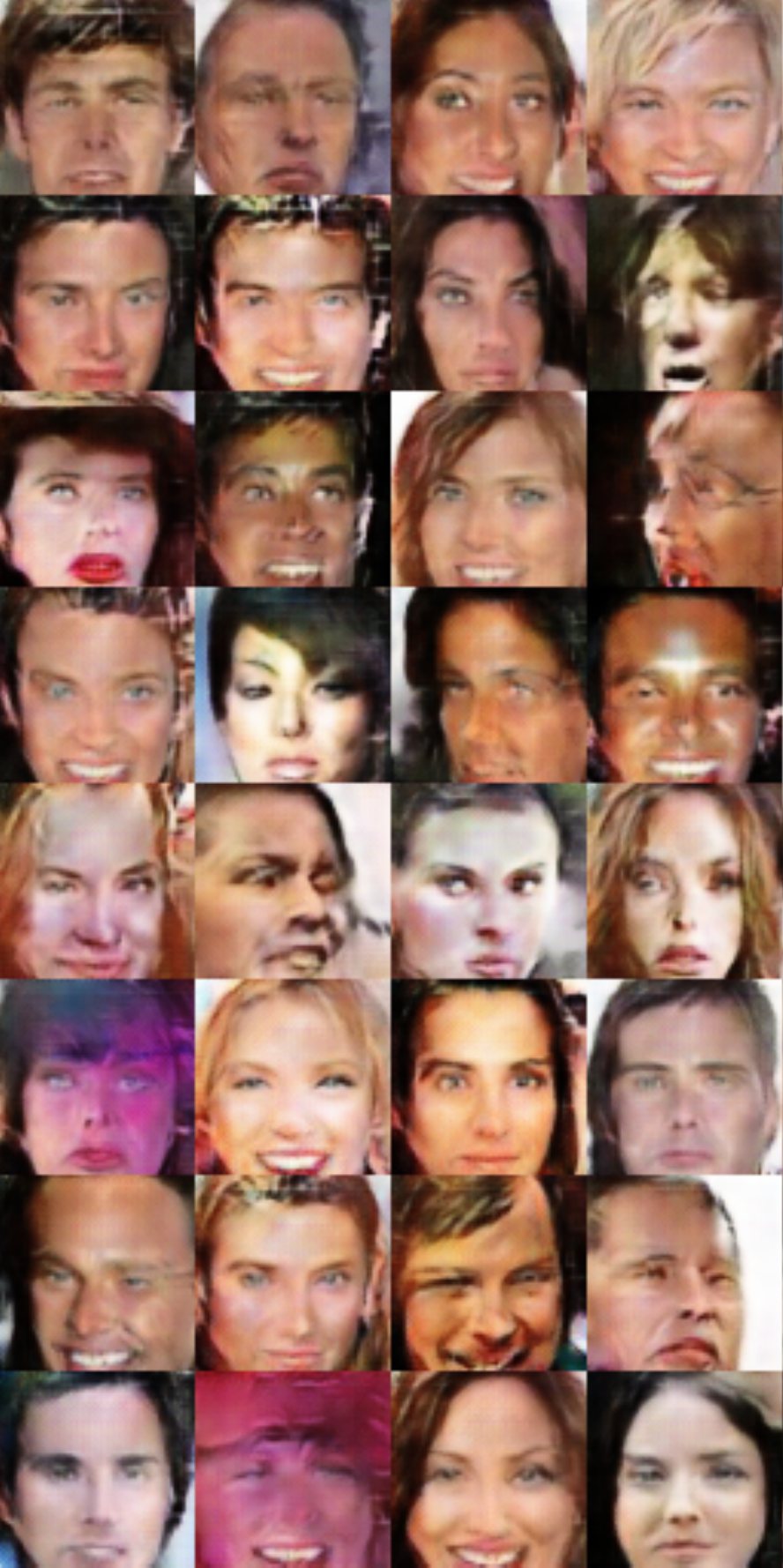}
	\caption{DCGAN, with BN}
\end{subfigure} ~
\begin{subfigure}{0.23\textwidth}
	\includegraphics[width=\textwidth]{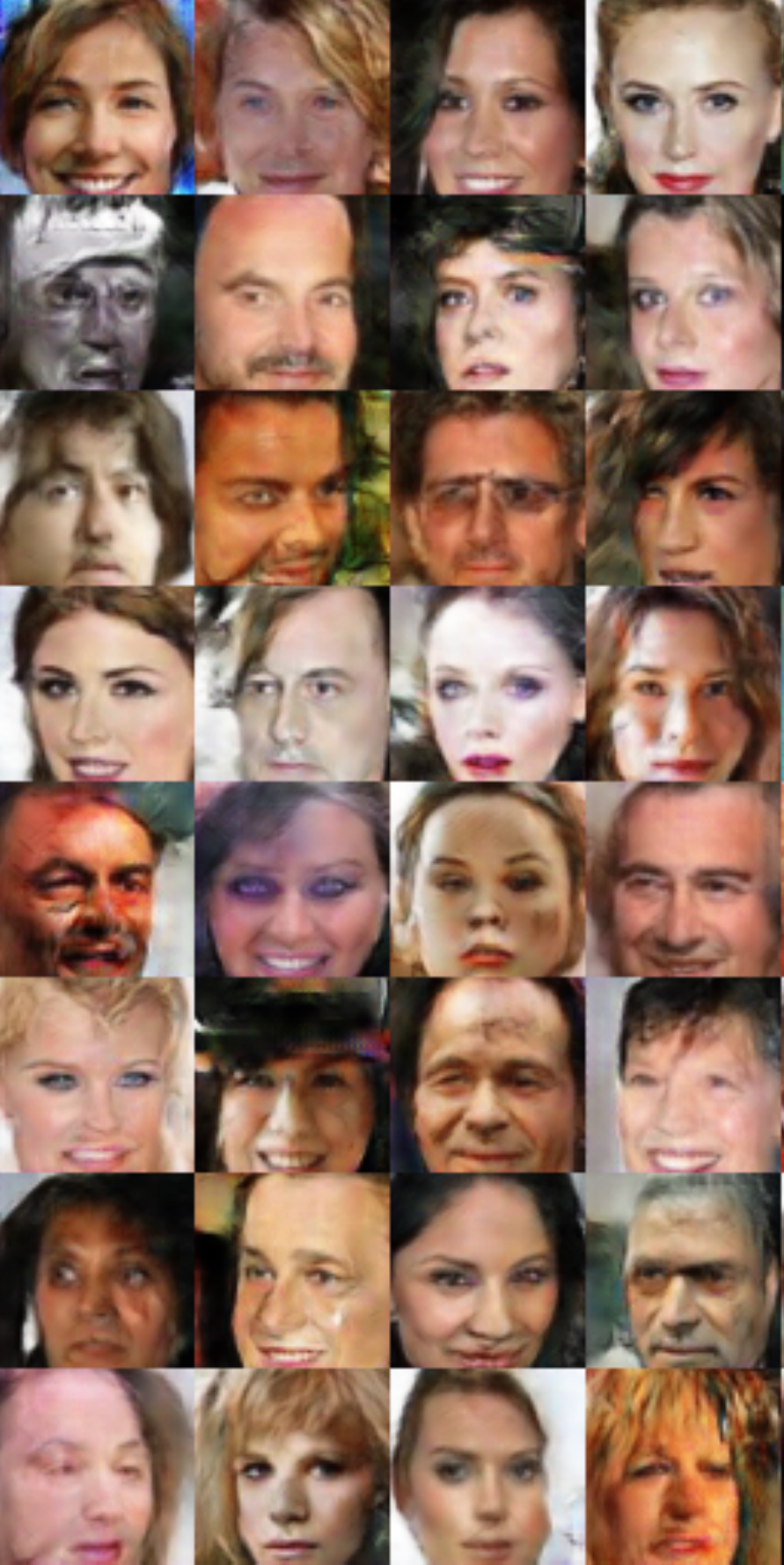}
	\caption{DAN-S, without BN}
\end{subfigure} ~
\begin{subfigure}{0.23\textwidth}
	\includegraphics[width=\textwidth]{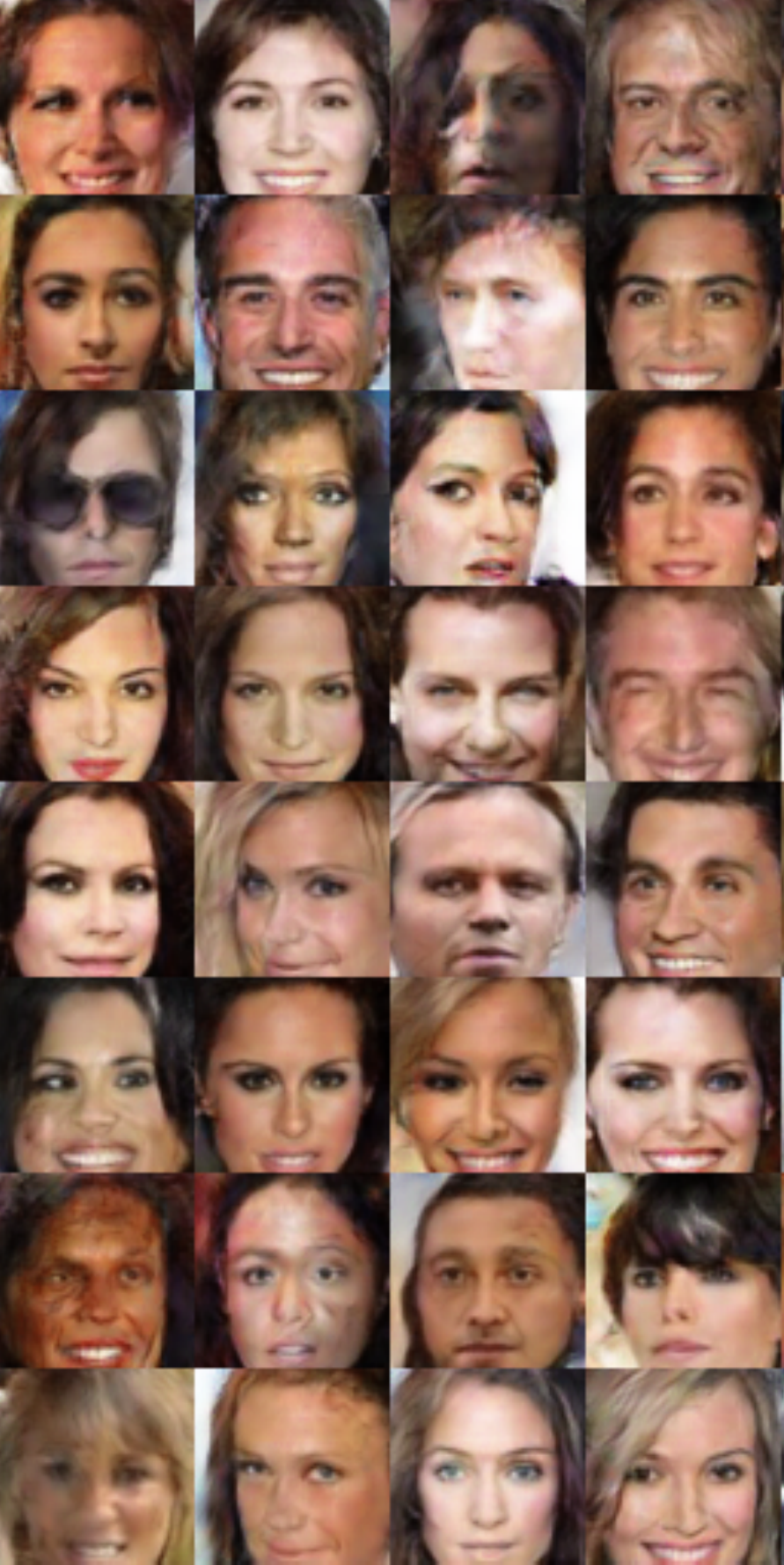}
	\caption{DAN-2S, without BN}
\end{subfigure}
\caption{GAN, DAN-S and DAN-2S trained on CelebA dataset to generate faces. }
\label{fig:celeba}
\end{figure}

We also test \ours on a harder image generation problem: generating faces. We train DCGAN and DAN with $\lambda_1 = 1$, $\lambda_2 = 0.2$ on the unlabeled CelebA~\cite{liu2015faceattributes} dataset and compare the generated samples. We use DCGAN with BN, and \ours-S/2S without BN. Figure~\ref{fig:celeba} visualizes half a minibatch (32 samples).

As discussed in Section~\ref{sec:relation}, both BN and \ours can be seen as techniques applying some form of normalization across a sample (minibatch). As seen in Figure~\ref{fig:celeba}, both DCGAN with BN and \ours-$\xi$ promotes diversity in generated samples as compared to DCGAN without BN. Such diverse samples indicate a better coverage of modes in the original dataset.

\subsection{Domain Adaptation}

For the last set of experiments, we test our \ours framework in the context of unsupervised domain adaptation, where the target domain data is fully unlabeled. This problem has been recently approached from the adversarial training perspective by using a domain classifier adversary to enforce domain-invariant representations~\cite{ganin2016domain, tzeng2017adversarial}. Following this approach, we use two neural network encoders (one for each domain) and a shared classifier. Enforcing the outputs of these encoders to be indistinguishable allows for a classifier trained on the source domain to be used in combination with the target encoder. For the adversary we use either \ours-S or \ours-2S to discern between samples of encoded representations from the two domains.

We first compare algorithms on the \textit{Amazon reviews} dataset preprocessed by~\citet{chen2012marginalized}. It consists of four domains: books, dvd, electronics and kitchen appliances, each of which contains reviews encoded in $5,000$ dimensional feature vectors and binary labels indicating if the review is positive or negative. We compare against the domain-adversarial neural network (DANN) of \citet{ganin2016domain}, which uses a similar framework but a single-sample-point discriminator. To make the comparison meaningful, we use the same architectures and parameters on all models, with the only difference being the representation averaging layer of our sample-based adversaries. We did not extensively tune parameters in any model. The results are shown in Table~\ref{tab:amazon}. Our models consistently outperform GAN-based DANN on most source-target pairs, with a considerable average improvement in accuracy of $1.41\%$ for \ours-S and $0.92\%$ for \ours-2S.
 %

\begin{table}[h!]
  \begin{center}
  	\begin{tabular}{llccc}
	\toprule
	Source & Target & DANN & \ours-S & \ours-2S \\
	\midrule
	books & dvd & 77.14(1.42) & 77.79(0.51) & 78.22(0.78) \\
	books & electronics & 74.38(1.12) & 75.68(0.42) & 74.78(1.08) \\
	books & kitchen & 77.23(1.09) & 79.01(0.41) & 76.90(0.58) \\
    dvd & books & 75.01(0.83) & 75.38(0.88) & 74.36(2.16) \\
    dvd & electronics & 75.62(0.85) & 75.98(0.91) & 76.28(1.37) \\
    dvd & kitchen & 79.41(0.96) & 80.37(0.91) & 79.96(0.89) \\
    electronics & books & 70.34(1.01) & 72.41(0.77) & 72.02(1.92) \\
    electronics & dvd & 69.42(2.60) & 72.39(1.99) & 72.34(1.65) \\
    electronics & kitchen & 83.64(0.48) & 85.02(0.37) & 84.66(0.68) \\
    kitchen & books & 69.54(0.88) & 70.39(0.85) & 70.45(1.41) \\
    kitchen & dvd & 69.36(2.24) & 73.85(0.82) & 71.50(2.93) \\
    kitchen & electronics & 82.68(0.47) & 82.47(0.16) & 83.38(0.28) \\
    \midrule
    avg. imp. && 0.00 & \textbf{1.41} & 0.92 \\
	\bottomrule
	\end{tabular}
  	\caption{Domain adaptation results on the Amazon review dataset. The numbers correspond to mean accuracies (plus one standard deviation) over 5 runs.}
	\label{tab:amazon}
  \end{center}
\end{table}

Lastly, we test the effectiveness of our proposed framework on a different domain adaptation task, this time for image label prediction. Specifically, we tackle the task of adapting from MNIST to MNIST-M, which is obtained by blending digits over patches randomly extracted from color photos from BSDS500~\cite{arbelaez2011contour} as in~\cite{ganin2016domain}. The results are shown in Table~\ref{tab:mnistm}. \ours-2S yields a strong improvement over DANN by $\sim3\%$. Again, the \ours-S and \ours-2S architectures consist of a simple plug-and-play adaptation of DANN to our distributional framework with little modifications, which demonstrates the ease of use of the proposed DAN criteria on top of existing models.

\begin{table}[h!]
  \begin{center}
  	\begin{tabular}{cc|cccc}
	\toprule
	Source Only & Target Only & DANN & \ours-S & \ours-2S \\
	\midrule
	52.20(1.94) & 95.60(0.16) & 76.13(1.63) & 77.07(1.94) & \textbf{79.214(2.19)} \\
	\bottomrule
	\end{tabular}
  	\caption{Results of domain adaptation from MNIST to MNIST-M, averaged over 5 different runs.}
	\label{tab:mnistm}
  \end{center}
\end{table}

\section{Discussion and Future Work}

The distributional adversarial framework we propose here is another step in an emerging trend of shifting from traditional classification losses to richer criteria better suited to finite-sample distribution discrimination. The experimental results obtained with this new approach offer a promising glimpse of the advantages of genuine sample-based discriminators over sample point alternatives, while the simplicity and ease of implementation makes this approach an appealing plug-in addition to existing models. 

Our framework is fairly general and opens the door to various possible extensions. The two types of discriminators proposed here are by no means the only options. There are many other approaches in the distributional discrepancy literature to draw inspiration from. One aspect that warrants additional investigation is the effect of sample size on training stability and mode coverage. It is sensible to expect that in order to maintain global discrimination power in settings with highly multimodal distributions, the size of samples fed to the discriminators should grow, at least with the number of modes. Formalizing this relationship is an interesting avenue for future work.

\bibliographystyle{abbrvnat}
\bibliography{refer}

\newpage

\begin{appendix}

\section{Training Algorithm for \ours}\label{app:sec:train}

We show the whole training procedure for \ours in Algorithm~\ref{app:algo:train}.

\begin{algorithm}[h!]
	\begin{algorithmic}
		\State\textbf{Input:}{total number of iterations $T$, size of minibatch $B$, step number $k$, model mode $\xi\in\{S,2S\}$}
		\For{$i=1$ to $T$}
			\State Sample minibatch $X=\{x^{(1)},\ldots,x^{(B)}\}\sim\bp_x$, $Z=\{z^{(1)},\ldots, z^{(B)}\}\sim\bp_z$
			\State Update discriminator by optimizing one step:
				\begin{align*}
				{\lambda_1\over B}\sum_{i=1}^B \left[\log (D(x^{(i)})) + \log (1 - D(G(z^{(i)})))\right]
				\end{align*}
			\If{mod($i$,$k$)=0}
				\State Sample minibatch $X$ and $Z$
				\If{$\xi = S$}
					\State Update $M_S$ by optimizing one step:
    					\begin{align*}
    					\lambda_2 (\log(M_S(\eta(X))) + \log(1 - M_S(\eta(G(Z)))))
    					\end{align*}
				\ElsIf{$\xi = 2S$}
					\State Divide $X$ and $Z$ into $X_1$, $X_2$ and $Z_1$, $Z_2$
					\State Update $M_{2S}$ by optimizing one step:
						\begin{align*}
						{\lambda_2\over 2} \{ &\log(M_{2S}(|\eta(X_1) - \eta(X_2)|)) +  \\
						&\log(M_{2S}(|\eta(G(Z_1)) - \eta(G(Z_2))|)) +\\
						&(1 - \log(M_{2S}(|\eta(X_1) - \eta(G(Z_2))|)) + \\
						&(1 - \log(M_{2S}(|\eta(G(Z_1)) - \eta(X_2)|))\}
						\end{align*}
				\EndIf
			\EndIf
			\State Sample minibatch $X$, $Z$
			\If{$\xi=S$}
				\State Update $G$ by optimizing one step:
					\begin{align*}
					{\lambda_1\over B}\sum_{i=1}^B \log (1 - D(G(z^{(i)}))) + \lambda_2 \log(1 - M_S(\eta(G(Z)))))
					\end{align*}
			\ElsIf{$\xi=2S$}
				\State Divide $X$ and $Z$ into $X_1$, $X_2$ and $Z_1$, $Z_2$
				\State Update $G$ by optimizing one step:
					\begin{align*}
					{\lambda_1\over B}&\sum_{i=1}^B \log (1 - D(G(z^{(i)}))) + \\
					&\lambda_2 (1 - \log(M_{2S}(|\eta(X_1) - \eta(G(Z_2))|)) + \\
						&(1 - \log(M_{2S}(|\eta(G(Z_1)) - \eta(X_2)|))\}
					\end{align*}
			\EndIf
		\EndFor
	\end{algorithmic}
	\caption{Training Procedure for \ours-S/2S.}\label{app:algo:train}
\end{algorithm}

\section{2D Gaussian Mixture Experiment Details} \label{app:sec:toy}

Following~\cite{metz2017unrolled}, we generate a mixture of $8$ Gaussian distribution with mean equally spaced on a circle of radius $2$ and variances of $0.01$. 

The generator consists of a fully connected network with 3 hidden layers of size 128 with \texttt{ReLU} activations, followed by a linear projection to 2 dimensions. The discriminator consists of a fully connected network with 3 hidden layers of size 32 with \texttt{ReLU} activations, followed by a linear projection to 1 dimension. Latent vectors are sampled uniformly from $[-1,1]^{256}$. Distributional adversaries starts with 2 hidden layers of size 32, then the latent representations are averaged across the batch. The mean representation is then fed to 1 hidden layer of size 32 and a linear projection to 1 dimension. For \ours we set $\lambda_1 = 0$ and $\lambda_2 = 1$.

Note that the number of parameters in GAN is the same as that in DAN.

We use Adam~\cite{kingma2014adam} with learning rate of 1E-4 and $\beta_1=0.5$ as training algorithm. Minibatch size is fixed to $512$ and we train the network for 25000 iterations by alternating between updates of generator and adversaries.

\section{MNIST Experiment Details} \label{app:sec:mnist}

We use the same architecture for generator across all models, which consists of 3 hidden layers with 256, 512 and 1,024 neurons each and with \texttt{ReLU} activations, followed by a fully connected linear layer to map it to 784 dimensional vector and finally a \texttt{sigmoid} activation for output. Latent vectors are uniform samples from $[-1,1]^{100}$. Decoder in EBGAN has the same architecture and encoders in RegGAN and EBGAN are the reverse of the decoder. For GAN, RegGAN and \ours the discriminator is the reverse of the generator except that the final linear function maps the vector of dimension 256 to 1. Distributional adversary in \ours has the same structure as the adversary in GAN but with an averaging of vectors across the batch before it is mapped to vector of dimension 256. And finally for RegGAN we set the two hyperparameters ($\lambda_1$ and $\lambda_2$ in the original paper~\cite{che2017mode}) to be $0.2$ and $0.4$, and for EBGAN we set the margin ($m$ in original paper~\cite{zhao2016energy}) to be $5$.

\begin{figure}[h!]
\centering
\begin{subfigure}{0.32\textwidth}
	\includegraphics[width=\textwidth]{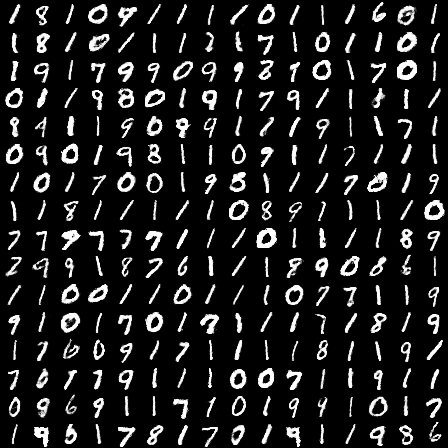}
	\caption{GAN}
\end{subfigure}	~
\begin{subfigure}{0.32\textwidth}
	\includegraphics[width=\textwidth]{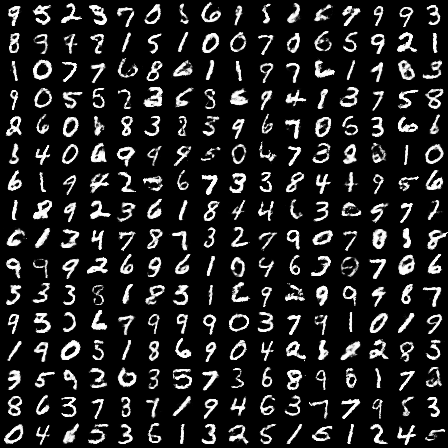}
	\caption{DAN-S}
\end{subfigure}
\begin{subfigure}{0.32\textwidth}
	\includegraphics[width=\textwidth]{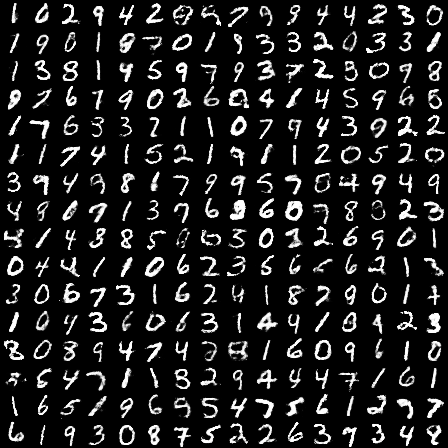}
	\caption{DAN-2S}
\end{subfigure} ~
\caption{GAN, DAN-S and DAN-2S trained on MNIST dataset to generate digits. DAN-S and DAN-2S clearly demonstrate better mode coverage and frequency recovery compared to GAN.}
\label{app:fig:mnist}
\end{figure}

We use Adam with learning rate of 5E-4 and $\beta_1=0.5$ for training. Minibatch size is fixed to 256 and we train the network for 50 epochs. For \ours we update the distributional adversaries every 5 iterations. For all other models we train by alternating between generator and adversaries (and potentially encoders and decoders).

We showcase the generated digits by GAN, DAN-S and DAN-2S in Figure~\ref{app:fig:mnist}, where we clearly see a better mode coverage of DAN-S and DAN-2S over GAN.

\section{CelebA Experiment Details} \label{app:sec:celeba}

We use public available code\footnote{\url{https://github.com/carpedm20/DCGAN-tensorflow}} for DCGAN. We set the network architecture the same as default in the code. The generator consists of a fully connected linear layer mapping from latent space of $[-1,1]^{100}$ to dimension 8,192, followed by 4 layers of deconvolution layers with \texttt{ReLU} activation except the last layer which is followed by \texttt{tanh}. The discriminator is the ``reverse'' of generator except that the activation function is \texttt{Leaky ReLU} and for the last layer, which is a linear mapping to 1 dimension and a \texttt{sigmoid} activation. \ours has the same architecture as DCGAN, except for the distributional adversary where it has one more layer of linear mapping with \texttt{ReLU} with 1,024 unites before the last linear mapping. We set $\lambda_1 = 1$ and $\lambda_2 = 0.2$.

All images are cropped to be $64\times 64$. We use Adam with learning rate of 2E-4 and $\beta_1=0.5$ for training. Batch size is fixed to 64 and we train the network for 20 epochs. We update the network by alternating between generator and adversaries. 

\begin{figure}[h!]
\centering

\begin{subfigure}{0.48\textwidth}
	\includegraphics[width=\textwidth]{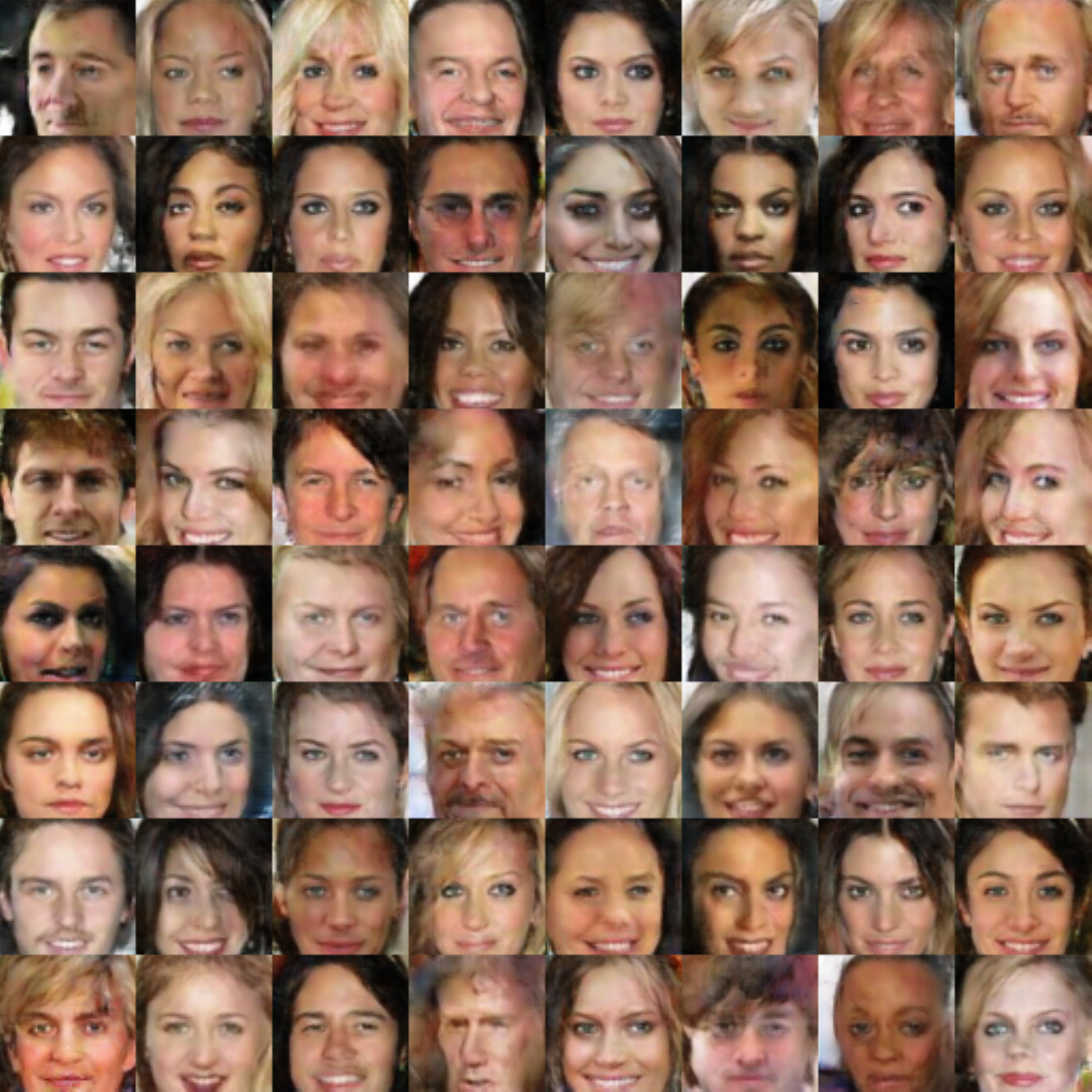}
	\caption{DCGAN, trained without BN}
\end{subfigure}	~
\begin{subfigure}{0.48\textwidth}
	\includegraphics[width=\textwidth]{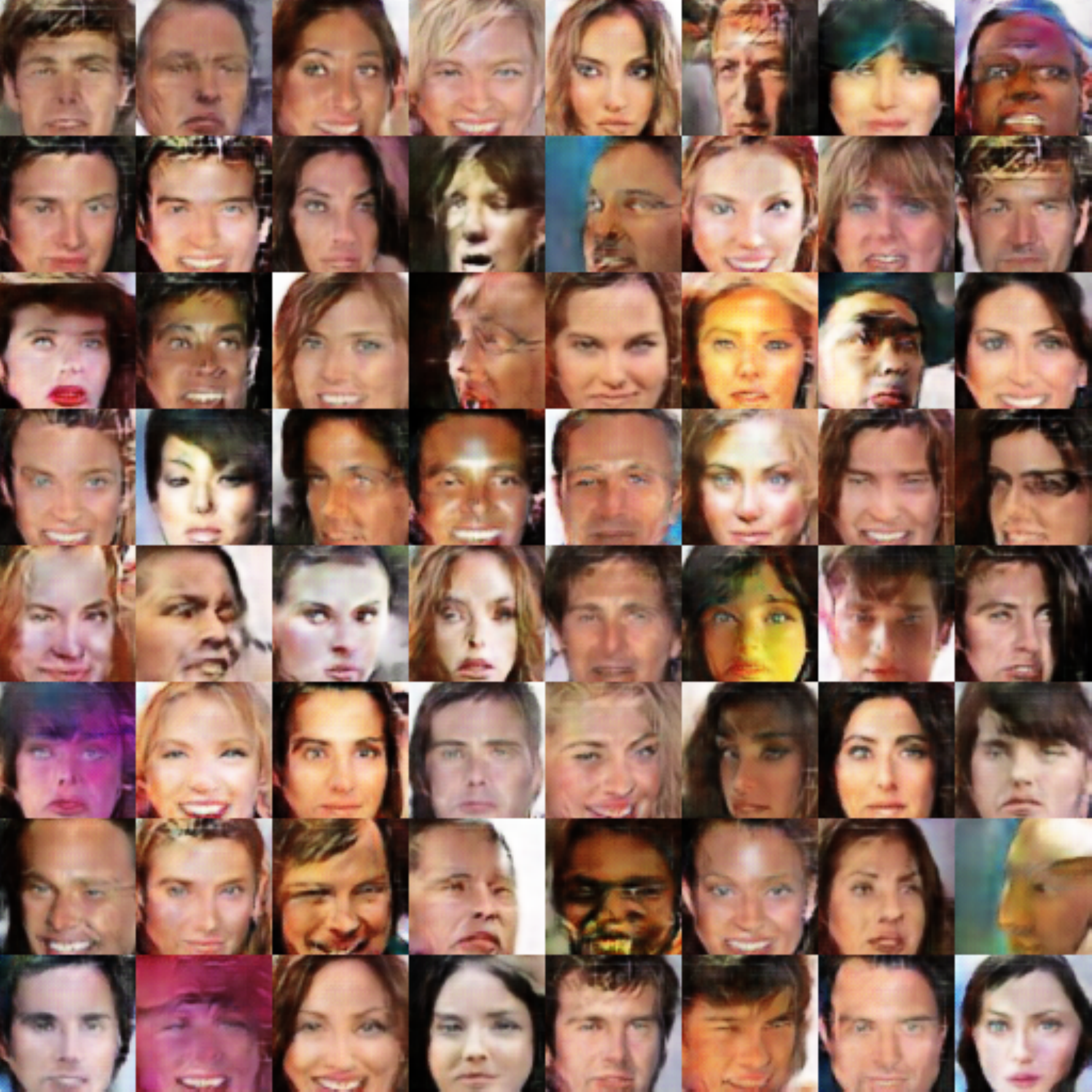}
	\caption{DCGAN, trained with BN}
\end{subfigure}
\begin{subfigure}{0.48\textwidth}
	\includegraphics[width=\textwidth]{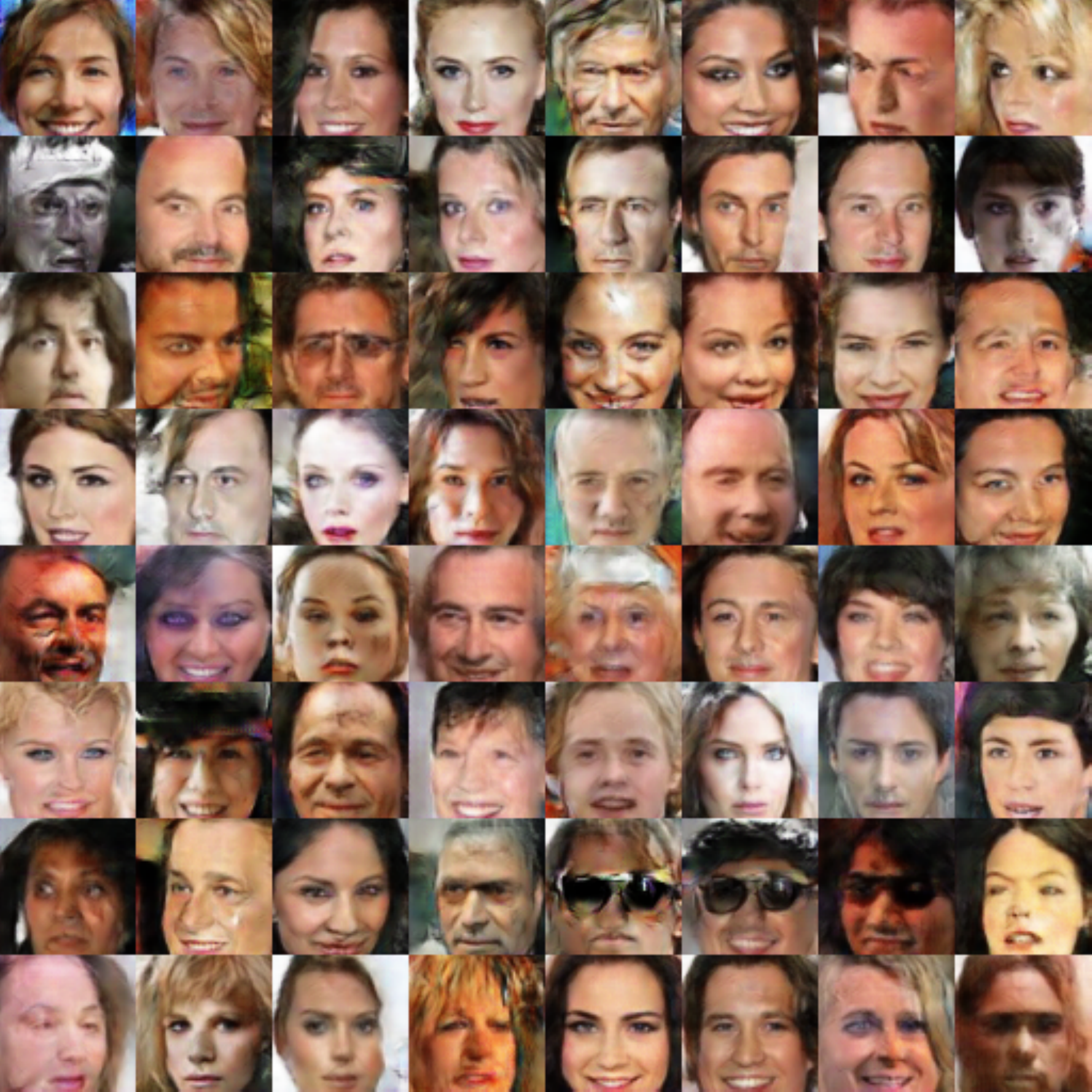}
	\caption{DAN-S, trained without BN}
\end{subfigure} ~
\begin{subfigure}{0.48\textwidth}
	\includegraphics[width=\textwidth]{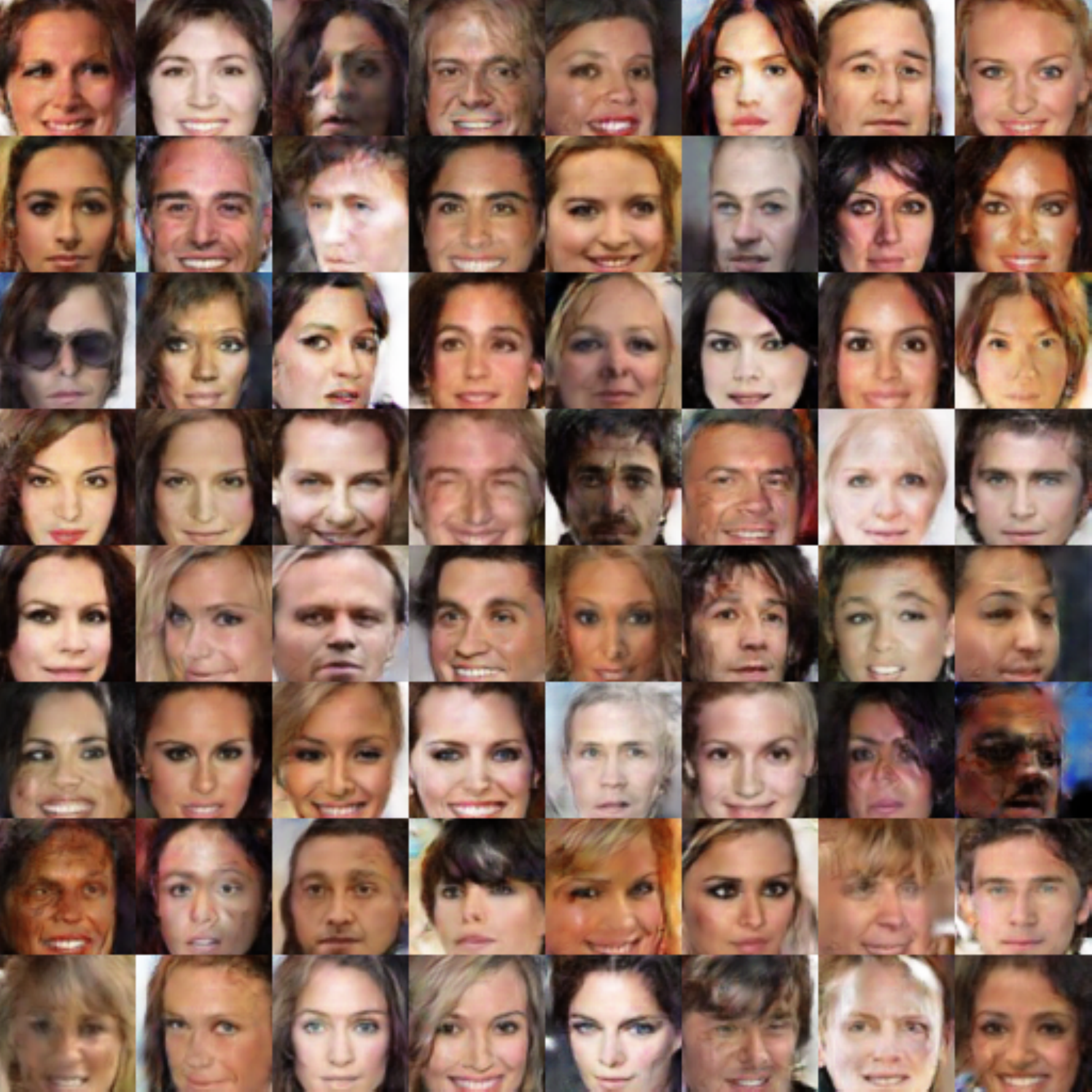}
	\caption{DAN-2S, trained without BN}
\end{subfigure}
\caption{GAN, DAN-S and DAN-2S trained on CelebA dataset to generate faces. DAN-S and DAN-2S demonstrate more diversity in generated samples compared to GAN trained without BN, and generates more realistic samples compared to that trained with BN.}
\label{app:fig:celeba}
\end{figure}

More samples from different models are shown in Figure~\ref{app:fig:celeba}.

\section{Domain Adaptation Experiment Details} \label{app:sec:da}

We adapt public available code\footnote{\url{https://github.com/pumpikano/tf-dann}} for DANN to train on Amazon and MNIST-M datasets. For the former dataset, we let the encoder for source/target domains consists of 3 fully connected linear layer mapping with \texttt{ReLU} activation with numbers of hidden units being $1000, 500, 100$ and is then mapped to 2-dimensional vector for both adversary and classifier. For the latter dataset we set the parameters to be the default values in the code. We set $\lambda_1=\lambda_2=0.1$ and $\lambda_1=\lambda_2=1$ for the two datasets respectively.

\end{appendix}

\end{document}